\documentclass{article}

\PassOptionsToPackage{round}{natbib}

\usepackage[final]{neurips_2019}

\usepackage[utf8]{inputenc} 
\usepackage[T1]{fontenc}    

\usepackage{hyperref}       
\hypersetup{
    colorlinks=true,
    linkcolor=black,      
    urlcolor=cyan,
    citecolor=black
}
 
\usepackage{url}            
\usepackage{booktabs}       
\usepackage{amsfonts}       
\usepackage{nicefrac}       
\usepackage{microtype}      

\usepackage{graphicx}
\usepackage{booktabs} 

\usepackage{amssymb}
\usepackage[ruled,lined]{algorithm2e}
\usepackage[lofdepth,lotdepth]{subfig}
\usepackage{amsthm}
\usepackage{color}
\usepackage{amsmath}
\usepackage{bigints}
\usepackage{multirow}
\usepackage{arydshln}
\usepackage{todonotes}

\let\d\relax
\DeclareMathOperator{\d}{d\!}

\title{MaCow: Masked Convolutional Generative Flow}

\author{Xuezhe Ma, Xiang Kong, Shanghang Zhang, Eduard Hovy \\ 
Carnegie Mellon University \\
Pittsburgh, PA, USA \\
\texttt{xuezhem,xiangk@cs.cmu.edu, shanghaz@andrew.cmu.edu, hovy@cmu.edu}
}

\begin{document}

\maketitle

\begin{abstract}
Flow-based generative models, conceptually attractive due to tractability of the exact log-likelihood computation and latent-variable inference as well as efficiency in training and sampling, has led to a number of impressive empirical successes and spawned many advanced variants and theoretical investigations.
Despite computational efficiency, the density estimation performance of flow-based generative models significantly falls behind those of state-of-the-art autoregressive models.
In this work, we introduce \emph{masked convolutional generative flow} (\textbf{\textsc{MaCow}}), a simple yet effective architecture for generative flow using masked convolution. 
By restricting the local connectivity to a small kernel, \textsc{MaCow} features fast and stable training along with efficient sampling while achieving significant improvements over Glow for density estimation on standard image benchmarks, considerably narrowing the gap with autoregressive models.
\end{abstract}

\section{Introduction}
Unsupervised learning of probabilistic models is a central yet challenging problem. 
Deep generative models have shown promising results in modeling complex distributions such as natural images~\citep{radford2015unsupervised}, audio~\citep{van2016wavenet} and text~\citep{bowman2015generating}.
Multiple approaches emerged in recent years, including Variational Autoencoders (VAEs)~\citep{kingma2014auto}, Generative Adversarial Networks (GANs)~\citep{goodfellow2014generative}, autoregressive neural networks~\citep{larochelle2011neural,oord2016pixel}, and flow-based generative models~\citep{dinh2014nice,dinh2016density,kingma2018glow}.
Among these, flow-based generative models gained popularity for this capability of estimating densities of complex distributions, efficiently generating high-fidelity syntheses, and automatically learning useful latent spaces.

Flow-based generative models typically warp a simple distribution into a complex one by mapping points from the simple distribution to the complex data distribution through a chain of invertible transformations with Jacobian determinants that are efficient to compute.
This design guarantees that the density of the transformed distribution can be analytically estimated, making maximum likelihood learning feasible.
Flow-based generative models have spawned significant interests for improving and analyzing its algorithms both theoretically and practically, and applying them to a wide range of tasks and domains.

In their pioneering work, \citet{dinh2014nice} first proposed \emph{Non-linear Independent Component Estimation}~(NICE) to apply flow-based models for modeling complex high-dimensional densities.
RealNVP~\citep{dinh2016density} extended NICE with a more flexible invertible transformation to experiment with natural images. 
However, these flow-based generative models resulted in worse density estimation performance compared to state-of-the-art autoregressive models, and are incapable of realistic synthesis of large images compared to GANs~\citep{karras2017progressive,brock2018large}.
Recently, \citet{kingma2018glow} proposed Glow as a generative flow with invertible $1\times1$ convolutions, which significantly improved the density estimation performance on natural images.
Importantly, they demonstrated that flow-based generative models optimized towards the plain likelihood-based objective are capable of generating realistic high-resolution natural images efficiently. 
\citet{prenger2018waveglow} investigated applying flow-based generative models to speech synthesis by combining Glow with WaveNet~\citep{van2016wavenet}.
\citet{ziegler2019latent} adopted variational inference to apply generative flows to discrete sequential data.
Unfortunately, the density estimation performance of Glow on natural images remains behind autoregressive models, such as PixelRNN/CNN~\citep{oord2016pixel,salimans2017pixelcnn++}, Image Transformer~\citep{parmar2018image}, PixelSNAIL~\citep{chen2017pixelsnail} and SPN~\citep{menick2018generating}.
There is also some work~\citep{rezende2015variational,kingma2016improved,zheng2017} trying to apply flow to variational inference. 

In this paper, we propose a novel architecture of generative flow, \emph{masked convolutional generative flow} (\textbf{\textsc{MaCow}}), which leverages masked convolutional neural networks~\citep{oord2016pixel}.
The bijective mapping between input and output variables is easily established while the computation of the determinant of the Jacobian remians efficient. 
Compared to inverse autoregressive flow~(IAF)~\citep{kingma2016improved}, \textsc{MaCow} offers stable training and efficient inference and synthesis by restricting the local connectivity in a small ``masked'' kernel as well as large receptive fields by stacking multiple layers of convolutional flows and using rotational ordering masks (\S\ref{subsec:macow}).
We also propose a fine-grained version of the multi-scale architecture adopted in previous flow-based generative models to further improve the performance (\S\ref{subsec:multi-scale}).
Experimenting with four benchmark datasets for images, CIFAR-10, ImageNet, LSUN, and CelebA-HQ, we demonstrate the effectiveness of \textsc{MaCow} as a density estimator by consistently achieving significant improvements over Glow on all the three datasets. 
When equipped with the variational dequantization mechanism~\citep{ho2019flow++}, \textsc{MaCow} considerably narrows the gap of the density estimation with autoregressive models (\S\ref{sec:experiment}).

\section{Flow-based Generative Models}\label{sec:background}
In this section, we first setup notations, describe flow-based generative models, and review Glow~\citep{kingma2018glow} as it is the foundation for \textsc{MaCow}.

\subsection{Notations}
Throughout the paper, uppercase letters represent random variables and lowercase letters for realizations of their corresponding random variables.
Let $X \in \mathcal{X}$ be the random variables of the observed data, e.g., $X$ is an image or a sentence for image and text generation, respectively.

Let $P$ denote the true distribution of the data, i.e., $X \sim P$, and $D = \{x_1, \ldots, x_N\}$ be our training sample, where $x_i, i=1,\ldots, N,$
are usually i.i.d.\ samples of $X$.
Let $\mathcal{P} = \{P_\theta : \theta \in \Theta\}$ denote a parametric statistical model indexed by the parameter $\theta \in \Theta$, where $\Theta$ is the parameter space.
$p$ denotes the density of the corresponding distribution $P$.
In the deep generative model literature, deep neural networks are the most widely used parametric models.
The goal of generative models is to learn the parameter $\theta$ such that $P_{\theta}$ can best approximate the true distribution $P$.
In the context of maximum likelihood estimation, we minimize the negative log-likelihood of the parameters with:
\begin{equation}\label{eq:mle}
\min\limits_{\theta \in \Theta} \frac{1}{N} \sum\limits_{i=1}^{N} -\log p_{\theta}(x_i) = \min\limits_{\theta \in \Theta} \mathrm{E}_{\widetilde{P}(X)} [-\log p_{\theta}(X)],
\end{equation}
where $\tilde{P}(X)$ is the empirical distribution derived from training data $D$.

\subsection{Flow-based Models}
In the framework of flow-based generative models, a set of latent variables $Z \in \mathcal{Z}$ are introduced with a prior distribution $p_{Z}(z)$, which is typically a simple distribution like a multivariate Gaussian.
For a bijection function $f: \mathcal{X} \rightarrow \mathcal{Z}$ (with $g = f^{-1}$), the change of the variable formula defines the model distribution on $X$ by
\begin{equation}
p_{\theta}(x) = p_{Z}\left(f_{\theta}(x)\right)\left| \det\left(\frac{\partial f_{\theta}(x)}{\partial x}\right)\right|,
\end{equation}
where $\frac{\partial f_{\theta}(x)}{\partial x}$ is the Jacobian of $f_{\theta}$ at $x$.

The generative process is defined straightforwardly as the following:
\begin{equation}
\begin{array}{rcl}
z & \sim & p_{Z}(z) \\
x & = & g_{\theta}(z).
\end{array}
\end{equation}
Flow-based generative models focus on certain types of transformations $f_{\theta}$ that allow the inverse functions $g_{\theta}$ and Jacobian determinants to be tractable to compute.
By stacking multiple such invertible transformations in a sequence, which is also called a (normalizing) \emph{flow}~\citep{rezende2015variational}, the flow is then capable of warping a simple distribution~($p_{Z}(z)$) into a complex one~($p(x)$) through:
\begin{displaymath}
X \underset{g_1}{\overset{f_1}{\longleftrightarrow}} H_1 \underset{g_2}{\overset{f_2}{\longleftrightarrow}} H2 \underset{g_3}{\overset{f_3}{\longleftrightarrow}} \cdots \underset{g_K}{\overset{f_K}{\longleftrightarrow}} Z,
\end{displaymath}
where $f = f_1 \circ f_2 \circ \cdots \circ f_K$ is a flow of $K$ transformations.
For brevity, we omit the parameter $\theta$ from $f_{\theta}$ and $g_{\theta}$.

\subsection{Glow}
Recently, several types of invertible transformations emerged to enhance the expressiveness of flows, among which Glow~\citep{kingma2018glow} has stood out for its simplicity and effectiveness on both density estimation and high-fidelity synthesis.
The following briefly describes the three types of transformations that comprise Glow.

\paragraph{Actnorm.} \citet{kingma2018glow} proposed an activation normalization layer (Actnorm) as an alternative for batch normalization~\citep{ioffe2015batch} to alleviate the challenges in model training.
Similar to batch normalization, Actnorm performs an affine transformation of the activations using a scale and bias parameter per channel for 2D images, such that
\begin{displaymath}
y_{i,j} = s \odot x_{i,j} + b,
\end{displaymath}
where both $x$ and $y$ are tensors of shape $[h\times w \times c]$ with spatial dimensions $(h, w)$ and channel dimension $c$.

\paragraph{Invertible $1 \times 1$ convolution.} To incorporate a permutation along the channel dimension, Glow includes a trainable invertible $1 \times 1$ convolution layer to generalize the permutation operation as:
\begin{displaymath}
y_{i, j} = W x_{i,j},
\end{displaymath}
where $W$ is the weight matrix with shape $c \times c$.

\paragraph{Affine Coupling Layers.} Following \citet{dinh2016density}, Glow includes affine coupling layers in its architecture of:
\begin{displaymath}
\begin{array}{rcl}
x_a, x_b & = & \mathrm{split}(x) \\
y_a & = & x_a \\
y_b & = & \mathrm{s}(x_a) \odot x_b + \mathrm{b}(x_a) \\
y & = & \mathrm{concat}(y_a, y_b),
\end{array}
\end{displaymath}
where $\mathrm{s}(x_a)$ and $\mathrm{b}(x_a)$ are outputs of two neural networks with $x_a$ as input.
The $\mathrm{split}()$ and $\mathrm{concat}()$ functions perform operations along the channel dimension.

From this designed architecture of Glow, we see that interactions between spatial dimensions are incorporated only in the coupling layers.
The coupling layer, however, is typically costly for memory resources, making it infeasible to stack a significant number of coupling layers into a single model, especially when processing high-resolution images.
The main goal of this work is to design a new type of transformation that simultaneously models the dependencies in both the spatial and channel dimensions while maintaining a relatively small memory footprint to improve the capacity of the generative flow.

\section{Masked Convolutional Generative Flows}\label{sec:macow}
In this section, we describe the architectural components of the \emph{masked convolutional generative flow} (\textsc{MaCow}). 
First, we introduce the proposed flow transformation using masked convolutions in \S\ref{subsec:macow}. 
Then, we present a fine-grained version of the multi-scale architecture adopted by previous generative flows~\citep{dinh2016density,kingma2018glow} in \S\ref{subsec:multi-scale}. 

\begin{figure}[t]
\centering
\includegraphics[width=0.7\textwidth]{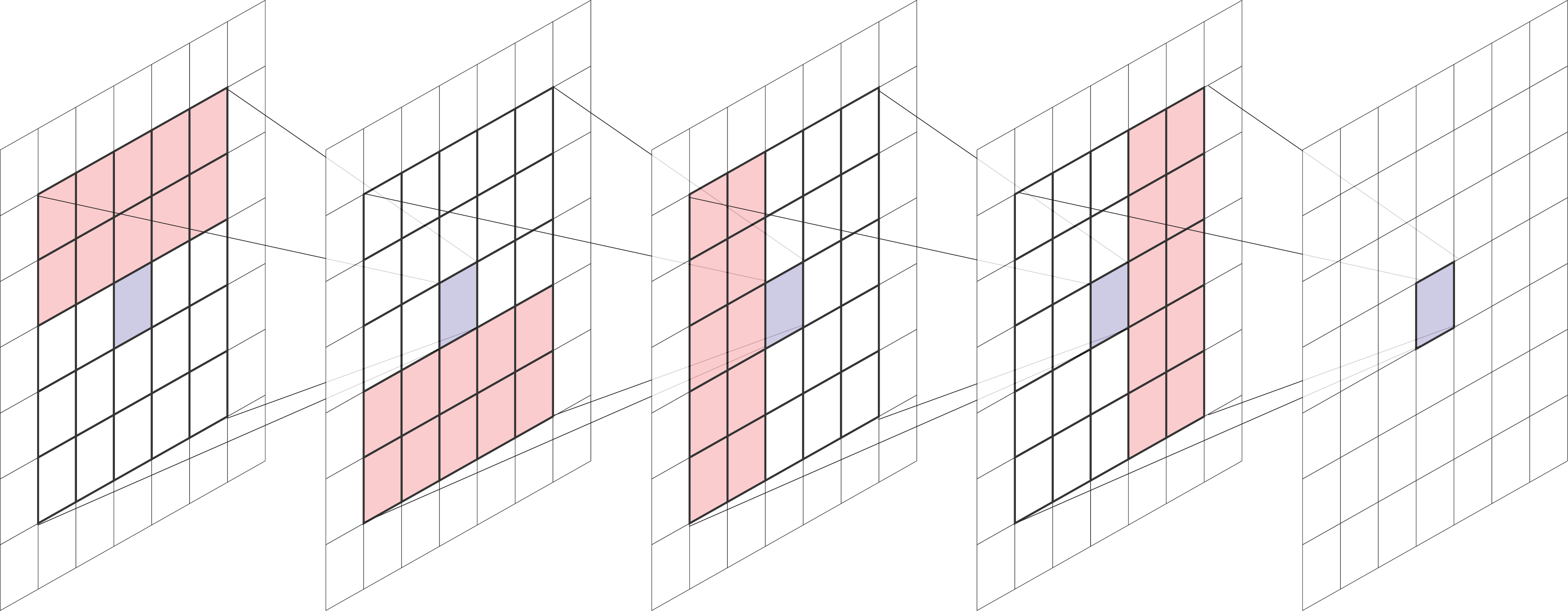}
\caption{Visualization of the receptive field of four masked convolutions with rotational ordering.}
\label{fig:mcf}
\end{figure}

\subsection{Flow with Masked Convolutions}\label{subsec:macow}
Applying autoregressive models to normalizing flows has been previously explored in studies~\citep{kingma2016improved,papamakarios2017masked}, with idea of sequentially modeling the input random variables in an autoregressive order to ensure the model cannot read input variables behind the current one:
\begin{equation}\label{eq:auto-flow}
y_t = \mathrm{s}(x_{<t}) \odot x_t + \mathrm{b}(x_{<t}),
\end{equation}
where $x_{<t}$ denotes the input variables in $x$ positioned ahead of $x_t$ in the autoregressive order.
$\mathrm{s}()$ and $\mathrm{b}()$ are two autoregressive neural networks typically implemented using spatial masks~\citep{germain2015made,oord2016pixel}.

Despite effectiveness in high-dimensional space, autoregressive flows suffer from two crucial problems: (1) The training procedure is unstable when stacking multiple layers to increase the flow capacities for complex distributions. (2) Inference and synthesis are inefficient, due to the non-parallelizable inverse function.

We propose to use masked convolutions to restrict the local connectivity in a small ``masked'' kernel to address these two problems.
The two autoregressive neural networks, $\mathrm{s}()$ and $\mathrm{b}()$, are implemented with one-layer masked convolutional networks with small kernels (e.g.\ $2\times5$ in Figure~\ref{fig:mcf}) to ensure they only read contexts in a small neighborhood based on:
\begin{equation}
\mathrm{s}(x_{<t}) = \mathrm{s}(x_{t\star}), \quad \mathrm{b}(x_{<t}) = \mathrm{b}(x_{t\star}),
\end{equation}
where $x_{t\star}$ denotes the input variables, restricted in a small kernel, on which $x_t$ depends.
By using masks in rotational ordering and stacking multiple layers of flows, the model captures a large receptive field (see Figure~\ref{fig:mcf}), and models dependencies in both the spatial and channel dimensions.

\paragraph{Efficient Synthesis.} As discussed above, synthesis from autoregressive flows is inefficient since the inverse must be computed by sequentially traversing through the autoregressive order. 
In the context of 2D images with shape $[h\times w \times c]$, the time complexity of synthesis is quadratic, i.e. $O(h \times w \times \mathrm{NN}(h, w, c))$, where $\mathrm{NN}(h, w, c)$ is the time of computing the outputs from the neural network $\mathrm{s}()$ and $\mathrm{b}()$ with input shape $[h\times w \times c]$. 
In our proposed flow with masked convolutions, computation of $x_{i,j}$ begins as soon as all $x_{t\star}$ are available, contrary to the autoregressive requirement that all $x_{<i,j}$ must have been already computed.
Moreover, at each step we only need to feed a slice of the image (with shape $[h\times kw\times c]$ or $[kh\times w\times c]$ depending on the direction of the mask) into $\mathrm{s}()$ and $\mathrm{b}()$.
Here $[\mathit{kh}\times \mathit{kw}\times c]$ is the shape of the kernel in the convolution.
Thus, the time complexity reduces significantly from quadratic to linear, which is $O(h \times \mathrm{NN}(\mathit{kh}, \mathit{w}, c))$ or $O(w \times \mathrm{NN}(\mathit{kw}, \mathit{h}, c))$ for horizontal and vertical masks, respectively.

\paragraph{Discussion}
The previous work closely related to \textsc{MaCow} is the Emerging Convolutions proposed in \citet{hoogeboom2019emerging}. 
There are two main differences.
i) the pattern of the mask is different. 
Emerging Convolutions use ``causal masks''~\citep{oord2016pixel} whose inverse function falls into a complete autoregressive transformation.
In contrast, \textsc{MaCow} achieves significantly more efficient inference and sampling (\S\ref{subsec:speed}), due to the carefully designed masks (Figure~\ref{fig:mcf}).
ii) the Emerging Convolutional Flow, proposed as an alternative to the invertible $1\times1$ convolution in Glow, is basically a linear transformation with masked convolutions, which does not introduce ``nonlinearity'' to the random variables.
\textsc{MaCow}, however, introduces such nonlinearity similar to the coupling layers. 

\begin{figure}[t]
\centering
\begin{minipage}[t]{0.99\linewidth}
\subfloat[One step of \textsc{MaCow}]{
\includegraphics[width=0.23\linewidth]{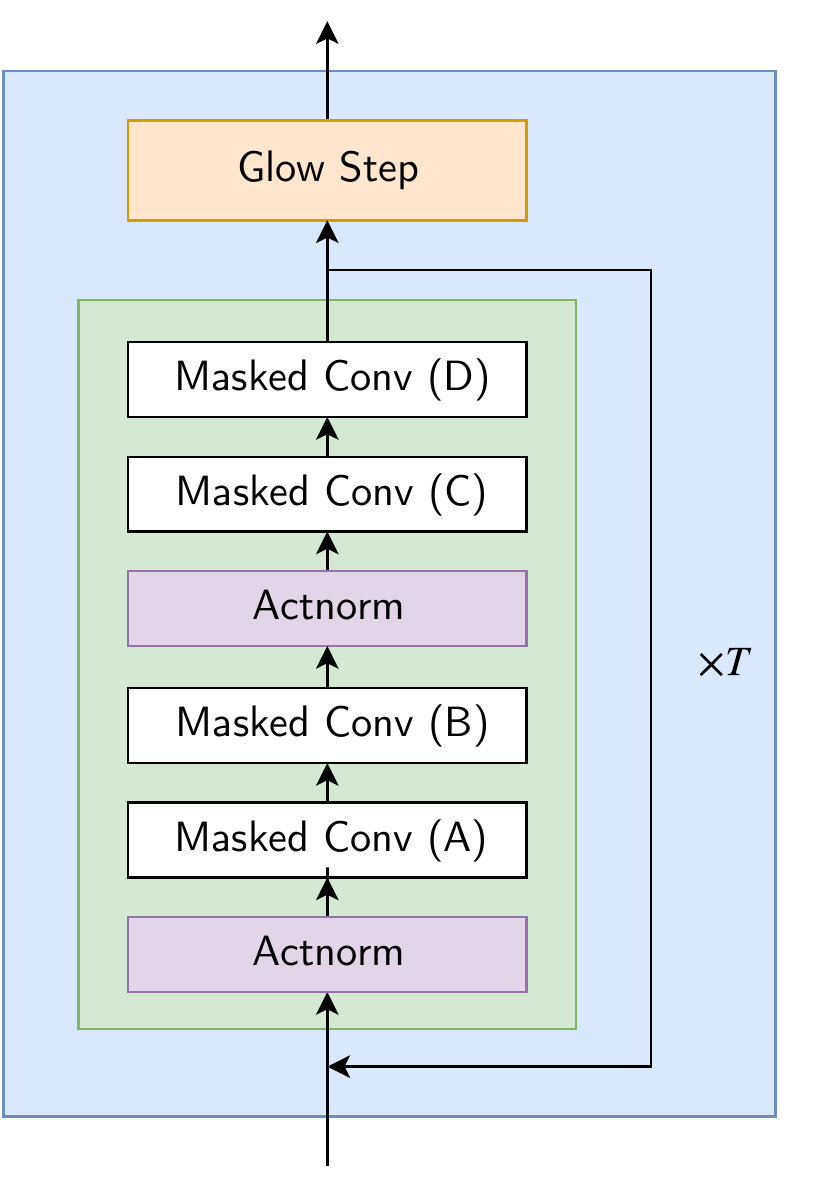}}
\label{fig:architecture:step}
\subfloat[Original multi-scale architecture]{
\includegraphics[width=0.34\linewidth]{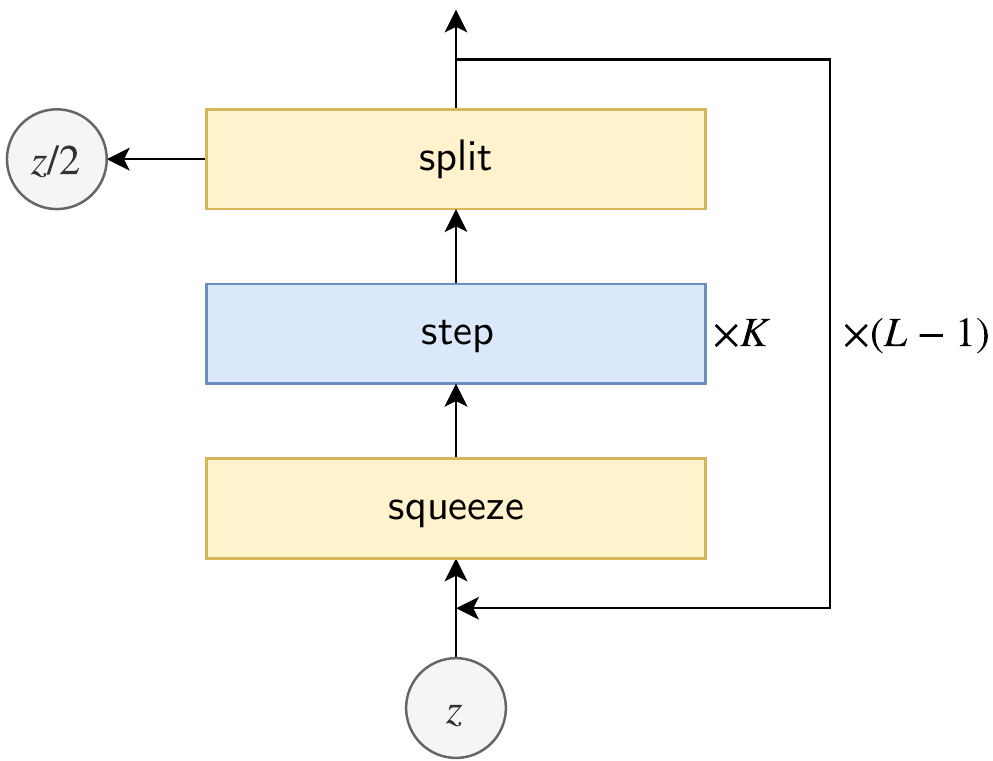}
\label{fig:architecture:multiscale}
}
\subfloat[Fine-grained multi-scale architecture]{
\includegraphics[width=0.39\linewidth]{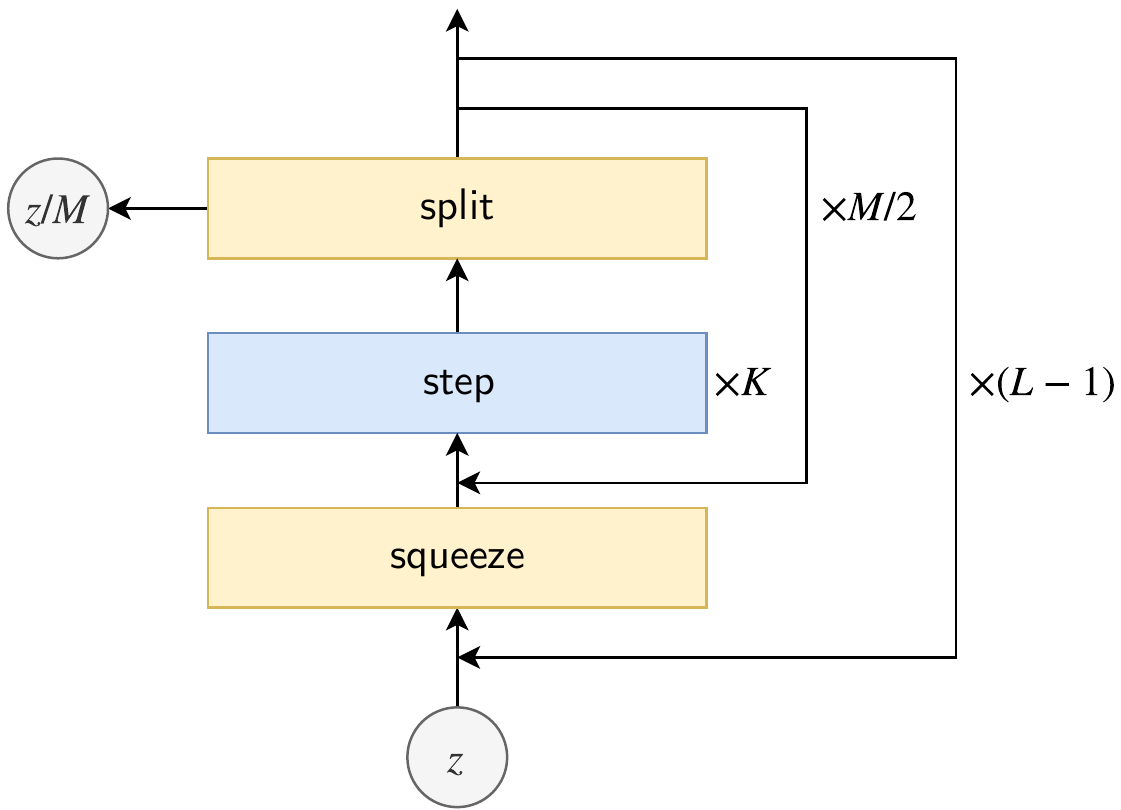}
\label{fig:architecture:fine-grained}
}
\end{minipage}
\caption{The architecture of the proposed \textsc{MaCow} model, where each step (a) consists of $T$ units of \emph{ActNorm} followed by two \emph{masked convolutions} with rotational ordering, and a Glow step. 
This flow is combined with either an original multi-scale (b) or a fine-grained architecture (c).}
\label{fig:architecture}
\end{figure}

\subsection{Fine-grained Multi-Scale Architecture}\label{subsec:multi-scale}
\citet{dinh2016density} proposed a multi-scale architecture using a squeezing operation, which has been demonstrated to be helpful for training very deep flows.
In the original multi-scale architecture, the model factors out half of the dimensions at each scale to reduce computational and memory costs.
In this paper, inspired by the size upscaling in subscale ordering~\citep{menick2018generating} that generates an image as a sequence of sub-images with equal size, we propose a fine-grained multi-scale architecture to improve model performance further.
In this fine-grained multi-scale architecture, each scale consists of $M/2$ blocks, and after each block, the model splits out $1/M$ dimensions of the input\footnote{In our experiments, we set $M=4$. Note that the original multi-scale architecture is a special case of the fine-grained version with $M=2$.}. 
Figure~\ref{fig:architecture} illustrates the graphical specification of the two architecture versions.
Note that the fine-grained architecture reduces the number of parameters compared with the original architecture with the same number of flow layers. 
Experimental improvements demonstrate the effectiveness of the fine-grained multi-scale architecture (\S\ref{sec:experiment}).

\section{Experiments}\label{sec:experiment}
We evaluate our \textsc{MaCow} model on both low-resolution and high-resolution datasets. 
For a step of \textsc{MaCow}, we use $T=2$ masked convolution units, and the Glow step is the same as that described in \citet{kingma2018glow} where an \emph{ActNorm} is followed by an \emph{Invertible $1\times1$ convolution}, which is followed by a \emph{coupling layer}. 
Each coupling layer includes three convolution layers where the first and last convolutions are $3\times3$, while the center convolution is $1\times1$.
For low-resolution images, we use affine coupling layers with 512 hidden channels, while for high-resolution images we use additive layers with 256 hidden channels to reduce memory cost.
ELU~\citep{clevert2015elu} is used as the activation function throughout the flow architecture.
For variational dequantization, the dequantization noise distribution $q_\phi(u|x)$ is modeled with a conditional \textsc{MaCow} with shallow architecture.
Additional details on architectures, results, and analysis of the conducted experiments are provided in Appendix~\ref{appendix:model}.

\subsection{Low-Resolution Images}
We begin our experiments with an evaluation of the density estimation performance of \textsc{MaCow} on two low-resolution image datasets that are commonly used to evaluate the deep generative models: CIFAR-10 with images of size $32\times32$~\citep{krizhevsky2009learning} and the $64\times64$ downsampled version of ImageNet~\citep{oord2016pixel}.

\begin{table}[t]
\caption{Density estimation performance on CIFAR-10 $32\times32$ and ImageNet $64\times64$. Results are reported in \emph{bits/dim}.}
\label{tab:density:low}
\centering
\begin{tabular}[t]{ll|c:c}
\toprule
& \textbf{Model} & \textbf{CIFAR-10} & \textbf{ImageNet-64} \\
\midrule
\multirow{8}{*}{Autoregressive} & IAF VAE~\citep{kingma2016improved} & 3.11 & -- \\
& Parallel Multiscale~\citep{reed2017parallel} & -- & 3.70 \\
& PixelRNN~\citep{oord2016pixel} & 3.00 & 3.63 \\
& Gated PixelCNN~\citep{van2016conditional} & 3.03 & 3.57 \\
& MAE~\citep{ma2019mae} & 2.95 & -- \\
& PixelCNN++~\citep{salimans2017pixelcnn++} & 2.92 & -- \\
& PixelSNAIL~\citep{chen2017pixelsnail} & \textbf{2.85} & \textbf{3.52} \\
& SPN~\citep{menick2018generating} & -- & \textbf{3.52} \\
\hline\midrule
\multirow{7}{*}{Flow-based} & Real NVP~\citep{dinh2016density} & 3.49 & 3.98 \\
& Glow~\citep{kingma2018glow} & 3.35 & 3.81 \\
& Flow++: \textsf{Unif}~\citep{ho2019flow++} & 3.29 & -- \\
& Flow++: \textsf{Var}~\citep{ho2019flow++} & \textbf{3.09} & 3.69 \\
\cline{2-4}
& \textsc{MaCow}: \textsf{Org} & 3.31 & 3.78 \\
& \textsc{MaCow}: \textsf{+fine-grained} & 3.28 & 3.75 \\
& \textsc{MaCow}: \textsf{+var} & 3.16 & \textbf{3.69} \\
\bottomrule
\end{tabular}
\end{table}

We perform experiments to dissect the effectiveness of each component of our \textsc{MaCow} model with ablation studies.
The \textsf{Org} model utilizes the original multi-scale architecture, while the \textsf{+fine-grained} model augments the original one with the fine-grained multi-scale architecture proposed in \S\ref{subsec:multi-scale}. 
The \textsf{+var} model further implements the variational dequantization on the top of \textsf{+fine-grained} to replace the uniform dequantization (see Appendix~\ref{appendix:dequant} for details).
For each ablation, we slightly adjust the number of steps in each level so that all the models have a similar number of parameters.

Table~\ref{tab:density:low} provides the density estimation performance for different variations of our \textsc{MaCow} model along with the top-performing autoregressive models (first section) and flow-based generative models (second section).
First, on both datasets, \textsf{fine-grained} models outperform \textsf{Org} ones, demonstrating the effectiveness of the fine-grained multi-scale architecture. 
Second, with the uniform dequantization, \textsc{MaCow} combined with the fine-grained multi-scale architecture significantly improves the performance over Glow on both datasets, and obtains slightly better results than Flow++ on CIFAR-10.
In addition, with variational dequantization, \textsc{MaCow} achieves comparable result in bits/dim with Flow++ on ImageNet $64\times64$.
On CIFAR-10, however, the performance of \text{MaCow} is around 0.07 bits/dim behind Flow++.

Compared with the state-of-the-art autoregressive generative models PixelSNAIL~\citep{chen2017pixelsnail} and SPN~\citep{menick2018generating}, the performance of \textsc{MaCow} is approximately 0.31 bits/dim worse on CIFAR-10 and 0.14 worse on ImageNet $64\times64$. 
Further improving the density estimation performance of \textsc{MaCow} on natural images is left to future work.

\begin{table}[t]
\caption{Negative log-likelihood scores for 5-bit LSUN and CelebA-HQ datasets in bits/dim.}
\label{tab:density:high}
\centering
\begin{tabular}[t]{l|c|c:c:c}
\toprule
 &  & \multicolumn{3}{c}{\textbf{LSUN}} \\
\textbf{Model} & \textbf{CelebA-HQ} & bedroom & tower & church \\
\toprule
Glow~\citep{kingma2018glow} & 1.03 & 1.20 & -- & --\\
SPN~\citep{menick2018generating} & \textbf{0.61} & -- & -- & -- \\
\midrule
\textsc{MaCow}: \textsf{Unif} & 0.95 & 1.16 & 1.22 & 1.36 \\
\textsc{MaCow}: \textsf{Var} & 0.67 & \textbf{0.98} & \textbf{1.02} & \textbf{1.09} \\
\bottomrule
\end{tabular}
\vspace{-1mm}
\end{table}

\begin{figure*}[t]
\centering
\begin{minipage}[t]{1.0\textwidth}
\begin{minipage}[t]{0.49\textwidth}
\subfloat[CelebA-HQ]{
\includegraphics[width=0.99\textwidth]{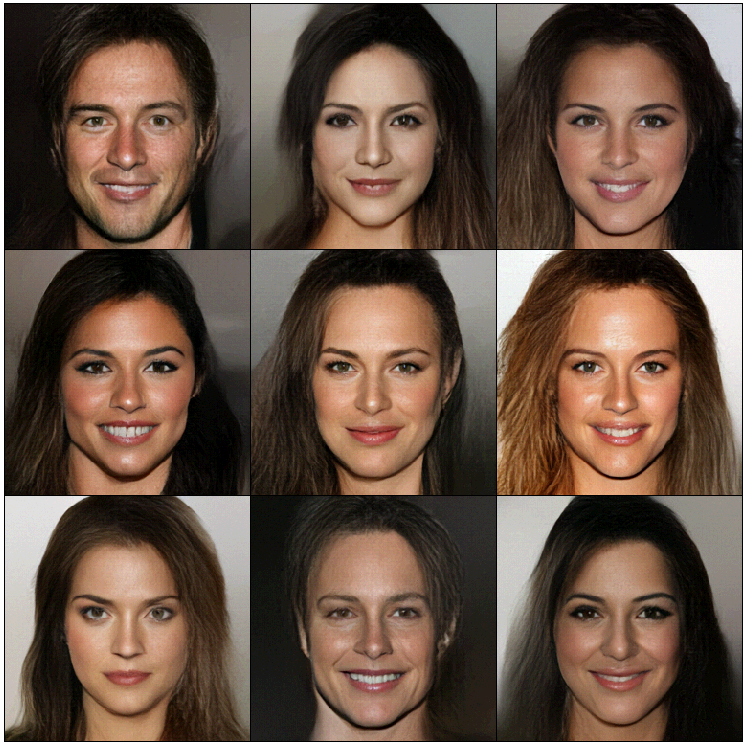}
}
\end{minipage}
\begin{minipage}[t]{0.49\textwidth}
\subfloat[LSUN church]{
\includegraphics[width=0.99\textwidth]{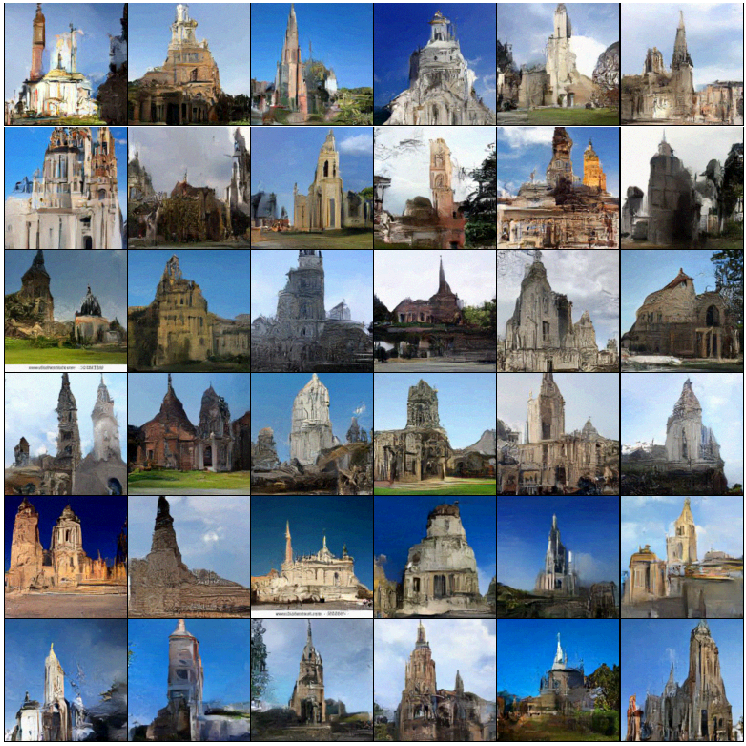}
}
\end{minipage}
\end{minipage}
\begin{minipage}[t]{1.0\textwidth}
\begin{minipage}[t]{0.49\textwidth}
\subfloat[LSUN tower]{
\includegraphics[width=0.99\textwidth]{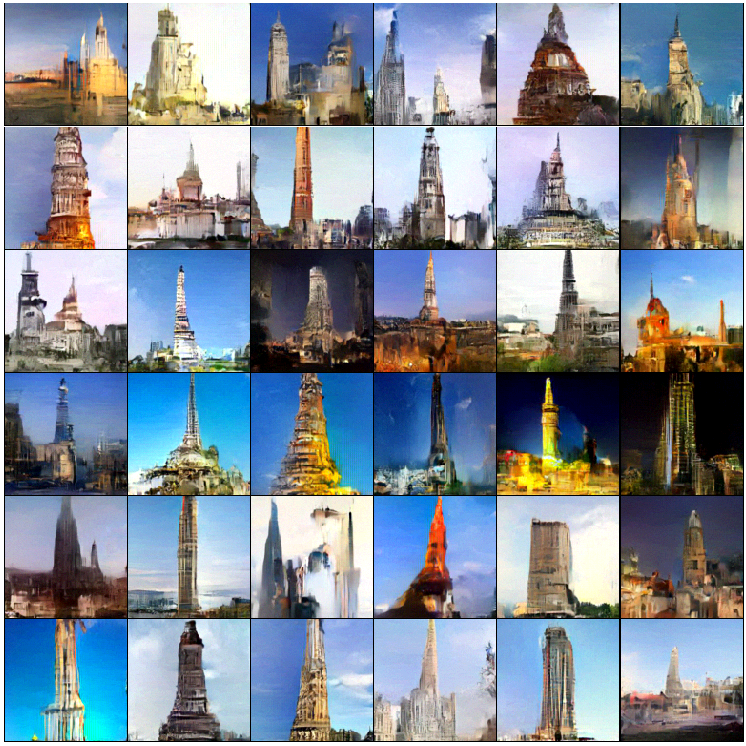}
}
\end{minipage}
\begin{minipage}[t]{0.49\textwidth}
\subfloat[LSUN bedroom]{
\includegraphics[width=0.99\textwidth]{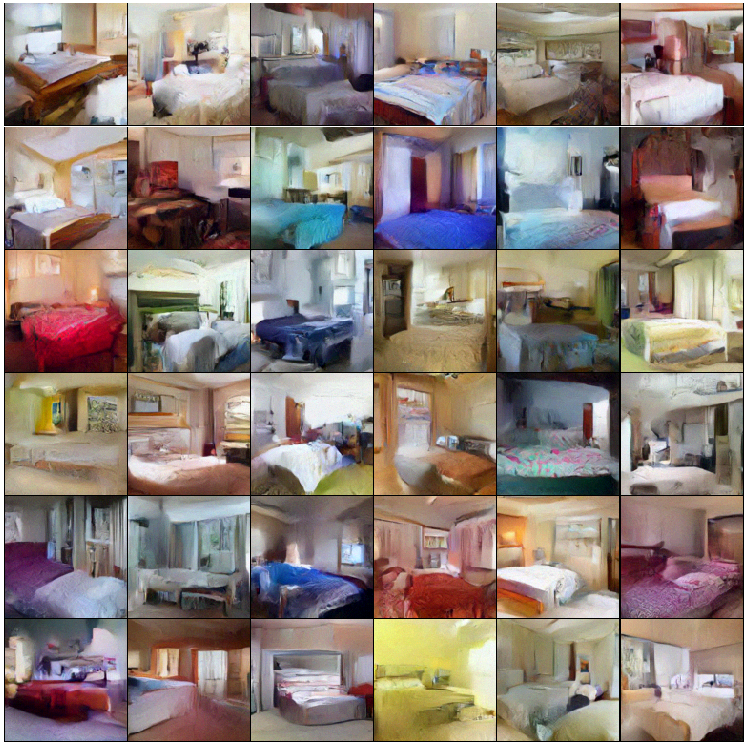}
}
\end{minipage}
\end{minipage}
\caption{(a) 5-bit $256\times256$ CelebA-HQ samples with temperature 0.7; (b)(c)(d) 5-bit $128\times128$ LSUN church, tower and bedroom samples, with temperature 0.9, respectively.}
\label{fig:celeba:sample}
\end{figure*}

\subsection{High-Resolution Images}
We next demonstrate experimentally that our \textsc{MaCow} model is capable of high fidelity samples at high-resolution.
Following \citet{kingma2018glow}, we choose the CelebA-HQ dataset~\citep{karras2017progressive}, which consists of 30,000 high-resolution images from the CelebA dataset~\citep{liu2015faceattributes}, and the LSUN~\citep{yu15lsun} datasets including categories \emph{bedroom}, \emph{tower} and \emph{church}.
We train our models on 5-bit images with the fine-grained multi-scale architecture and both the uniform and variational dequantization.
For each model, we adjust the number of steps in each level so that all the models have similar numbers of parameters with Glow for a fair comparison.

\subsubsection{Density Estimation}
Table~\ref{tab:density:high} illustrates the negative log-likelihood scores in bits/dim of two versions of \textsc{MaCow} on the 5-bit $128\times128$ LSUN and $256\times256$ CelebA-HQ datasets.
With uniform dequantization, \textsc{MaCow} improves the log-likelihood over Glow from 1.03 bits/dim to 0.95 bits/dim on CelebA-HQ dataset. 
Equipped with variational dequantization, \textsc{MaCow} obtains 0.67 bits/dim, which is 0.06 bits/dim behind the state-of-the-art autoregressive generative model SPN~\citep{menick2018generating} and significantly narrows the gap. 
On the LSUN datasets, \textsc{MaCow} with uniform dequantization outperforms Glow with 0.4 bits/dim on the bedroom category.
With variational dequantization, the model achieves further improvements on all the three categories of LSUN datasets, 

\subsubsection{Image Generation}
Consistent with previous work on likelihood-based generative models~\citep{parmar2018image,kingma2018glow}, we found that sampling from a reduced-temperature model often results in higher-quality samples.
Figure~\ref{fig:celeba:sample} showcases some random samples for 5-bit CelebA-HQ $256\times256$ with temperature 0.7 and LSUN $128\times128$ with temperature 0.9.
The images are extremely high quality for non-autoregressive likelihood models, despite that maximum likelihood is a principle that values diversity over sample quality in a limited capacity setting~\citep{theis2016note}.
More samples of images, including samples of low-resolution ones, are provided in Appendix~\ref{appendix:samples}\footnote{The reduced-temperature sampling is only applied to LSUN and CelebA-HQ 5-bits images, where \textsc{MaCow} adopts additive coupling layers. For CIFAR-10 and ImageNet 8-bits images, we sample with temperature 1.0.}.

\begin{table}[t]
\caption{(a) Image synthesis speed on CIFAR10. Glow re-implemented in PyTorch is masked with $\dag$. $\ddag$ denotes results shown in ~\citet{hoogeboom2019emerging}. (b) Image synthesis speed of \textsc{MaCow} on datasets with different image sizes. The time is measured in milliseconds to sample a datapoint when computed in mini-batchs with size $100$.}
\label{tab:speed}
\begin{minipage}[t]{0.49\textwidth}
\centering
\subfloat[]{
\label{tab:cifar10}
\begin{tabular}{l|c|c}
\toprule
    CIFAR10 & time (ms) & Slow-down\\
\midrule
    Glow$^\ddag$     & 5  & 1.0 \\
    MAF $^\ddag$     & 3000 & 600.0 \\
    Emerging$^\ddag$ & 1800 & 360.0 \\
\midrule
    Glow$^\dag$ & 5.3 & 1.0 \\
    \textsc{MaCow} & 38.7 & 7.3 \\
\bottomrule
\end{tabular}}
\end{minipage}
\hfill
\begin{minipage}[t]{0.49\textwidth}
\centering
\subfloat[]{
\label{tab:high}
\begin{tabular}{l|c|c}
\toprule
    Dataset & image size & time (ms) \\
\midrule
    CIFAR10 & $32 \times 32$ & 38.7 \\
    ImageNet & $64 \times 64$  & 104.7 \\
    LSUN   & $128 \times 128$   & 267.9 \\
    CelebA-HQ & $256 \times 256$ & 434.2 \\
\bottomrule
\end{tabular}}
\end{minipage}
\end{table}

\subsection{Comparison on Synthesis Speed}\label{subsec:speed}
In this section, we compare the synthesis speed of \textsc{MaCow} at test time with that of Glow~\citep{kingma2018glow}, Masked Autoregressive Flows (MAF)~\citep{papamakarios2017masked} and Emerging Convolutions~\citep{hoogeboom2019emerging}.
Following \citet{hoogeboom2019emerging}, we measure the time to sample a datapoint when computed in mini-batchs with size $100$. 
For fair comparison, we re-implemented Glow using PyTorch~\citep{paszke2017automatic}, and all experiments are conducted on a single NVIDIA TITAN X GPU.

Table~\ref{tab:cifar10} shows the sampling speed of \textsc{MaCow} on CIFAR-10, together with that of the baselines. 
\textsc{MaCow} is 7.3 times slower than Glow, much faster than Emerging Convolution and MAF, whose factors are 360 and 600 respectively.
The sampling speed of \textsc{MaCow} on datasets with different image sizes is shown in Table~\ref{tab:high}.
We see that the time of synthesis increases approximately linearly with the increase of image resolution.

\section{Conclusion}
In this paper, we propose a new type of generative flow, coined \textsc{MaCow}, which exploits masked convolutional neural networks.
By restricting the local dependencies in a small masked kernel, \textsc{MaCow} boasts fast and stable training as well as efficient sampling.
Experiments on both low- and high-resolution benchmark datasets of images show the capability of \textsc{MaCow} on both density estimation and high-fidelity generation, achieving state-of-the-art or comparable likelihood as well as its superior quality of samples compared to previous top-performing models\footnote{https://github.com/XuezheMax/macow}

A potential direction for future work is to extend \textsc{MaCow} to other forms of data, in particular text, on which no attempt (to the best of our knowledge) has been made to apply flow-based generative models.
Another exciting direction is to combine \textsc{MaCow} with variational inference to automatically learn meaningful (low-dimensional) representations from raw data.

\bibliography{macow}
\bibliographystyle{plainnat}

\newpage
\section*{Appendix: MaCow: Masked Convolutional Generative Flow}
\appendix
\setcounter{equation}{0}

\section{Dequantization}\label{appendix:dequant}
As described in \S\ref{sec:background}, generative flows are defined on continuous random variables.
Many real-world datasets, however, are recordings of discrete representations of signals, and fitting a continuous density model to discrete data produces a degenerate solution that places all probability mass on discrete datapoints~\citep{uria2013rnade,ho2019flow++}.
A common solution to this problem is ``dequantization'' that converts the discrete data distribution into a continuous one.

Specifically, in the context of natural images, each dimension (pixel) of the discrete data $x$ takes on values in $\{0, 1, \ldots, 255\}$.
The dequatization process adds continuous random noise $u$ to $x$, resulting a continuous data point of:
\begin{equation}\label{eq:dequant}
y = x + u,
\end{equation}
where $u \in [0, 1)^{d}$ is continuous random noise taking values from interval $[0, 1)$.
By modeling the density of $Y \in \mathcal{Y}$ with $p_{\theta}(y)$, the distribution of $X$ is defined as:
\begin{equation}
P_\theta(x) = \int_{\mathcal{Y}} p_\theta(y) \d y = \int_{[0, 1)^d} p_\theta(x + u) \d u.
\end{equation}
By restricting the range of $u$ in $[0, 1)$, the mapping between $y$ and a pair of $x$ and $u$ is bijective. Thus, we have $p_\theta(y) = p_\theta(x + u) = p_\theta(x, u)$.

By introducing a \emph{dequantization noise distribution} $q(u|x)$, the training objective in \eqref{eq:mle} can be re-written as:
\begin{align}
\mathrm{E}_{P(X)} \Big[-\log P_{\theta}(X)\Big] & = \mathrm{E}_{P(X)} \left[-\log \int_{[0, 1)^d} p_\theta(X, u) \d u \right] \nonumber \\
 & = \mathrm{E}_{P(X)} \Bigg[\mathrm{E}_{q(u|X)} \left[-\log \frac{p_{\theta}(X, u)}{q(u|X)}\right] - \mathrm{KL}\big(q(u|X) || p_{\theta}(u|X)\big) \Bigg] \nonumber \\
 & \leq \mathrm{E}_{P(X)} \Bigg[\mathrm{E}_{q(u|X)} \Big[-\log p_{\theta}(X, u) \Big] + \mathrm{E}_{q(u|X)} \Big[\log q(u|X) \Big]\Bigg] \nonumber \\
 & = \mathrm{E}_{p(Y)} \Big[-\log p_{\theta}(Y) \Big] + \mathrm{E}_{P(X)} \mathrm{E}_{q(u|X)} \Big[\log q(u|X)\Big], \label{eq:elbo}
\end{align}
where $p(y) = P(x) q(u|x)$ is the distribution of the dequantized variable $Y$ under the dequantization noise distribution $q(u|X)$.

\paragraph{Uniform Dequantization.} The most common dequantization method in prior work is uniform dequantization where the noise $u$ is sampled from the uniform distribution $\mathrm{Unif(0, 1)}$ such that
\begin{displaymath}
q(u|x) \sim \mathrm{Unif(0, 1)}, \forall x \in \mathcal{X}.
\end{displaymath}
From \eqref{eq:elbo}, we have 
\begin{displaymath}
\mathrm{E}_{P(X)} \left[-\log P_{\theta}(X)\right] \leq \mathrm{E}_{p(Y)} \left[-\log p_{\theta}(Y) \right],
\end{displaymath}
as $\log q(u|x) = 0, \forall x \in \mathcal{X}$.

\paragraph{Variational Dequantization.} As discussed in \citet{ho2019flow++}, uniform dequantization directs $p_\theta(y)$ to assign uniform density to unit hypercubes $[0, 1)^d$, which is difficult for smooth distribution approximators.
They proposed a parametric dequantization noise distribution $q_\phi(u|x)$ with a training objective to optimize the \emph{evidence lower bound} (ELBO) provided in \eqref{eq:elbo}:
\begin{equation}
\min\limits_{\theta, \phi} \mathrm{E}_{p_\phi(Y)} \left[-\log p_{\theta}(Y) \right] + \mathrm{E}_{P(X)} \mathrm{E}_{q_\phi(u|X)} \left[\log q_\phi(u|X) \right],
\end{equation}
where $p_\phi(y) = P(x) q_\phi(u|x)$.
In this paper, we implemented both these two dequantization methods for our \textsc{MaCow}, as is detailed in \S\ref{sec:experiment}).

\newpage
\section{Experimental Details}\label{appendix:model}
\subsection{Model details}
 
\begin{table}[h]
\caption{Hyper-parameters for \textsc{MaCow} in our experiments.}
\label{tab:my_label}
\centering
\resizebox{1.0\columnwidth}{!}
{
\begin{tabular}[t]{c|c|c|c|c|c|c}
\toprule
DataSet &  Dequant & Batch Size & Levels & Depths per Level & \# Param & \# Param Glow \\
\midrule
\multirow{2}{*}{CIFAR-10} & Unif & 512 & 3 & $[[12, 12], [12, 12], 12]$ & 41.2M & \multirow{2}{*}{44.2M} \\
 & Var  & 512 & 3 & $[[12, 12], [12, 12], 12]$ & 43.5M &  \\
\midrule
\multirow{2}{*}{ImageNet} & Unif & 160 & 4 & $[[16, 16], [16, 16], [12, 12], 12]$ & 117.2M & \multirow{2}{*}{111.6M} \\
 & Var & 160 & 4 & $[[16, 16], [16, 16], [12, 12], 12]$ & 122.5M &  \\
\midrule
\multirow{2}{*}{LSUN} & Unif & 160 & 5 & $[[32, 32], [32, 32], [16, 16], [12, 12], 6]$ & 166.6M & \multirow{2}{*}{198.1M} \\
 & Var & 160 & 5 & $[[32, 32], [32, 32], [16, 16], [12, 12], 6]$ & 171.9M & \\
\midrule
\multirow{2}{*}{CelebA-HQ} & Unif & 40 & 6 & $[[24, 24], [16, 16], [16, 16], [8, 8], [4, 4], 2]$ & 171.9M & \multirow{2}{*}{170.8M} \\
 & Var & 40 & 6 & $[[24, 24], [16, 16], [16, 16], [8, 8], [4, 4], 2]$ & 177.3M & \\
\bottomrule
\end{tabular}
}
\end{table}

\subsection{Optimization}
Parameter optimization is performed with the Adam optimizer~\citep{kingma2014adam} with $\beta=(0.9, 0.999)$ and $\epsilon=1\mathrm{e}-8$. 
Warmup training is applied to all the experiments: the learning rate linearly increases to  for 500 updates to the initial learning rate $1\mathrm{e}-3$.
Then we use exponential decay to decrease the learning rate with decay rate is $0.999997$.

\newpage
\section{More samples from our experiments}\label{appendix:samples}

\begin{figure}[h]
    \centering
    \includegraphics[width=\linewidth]{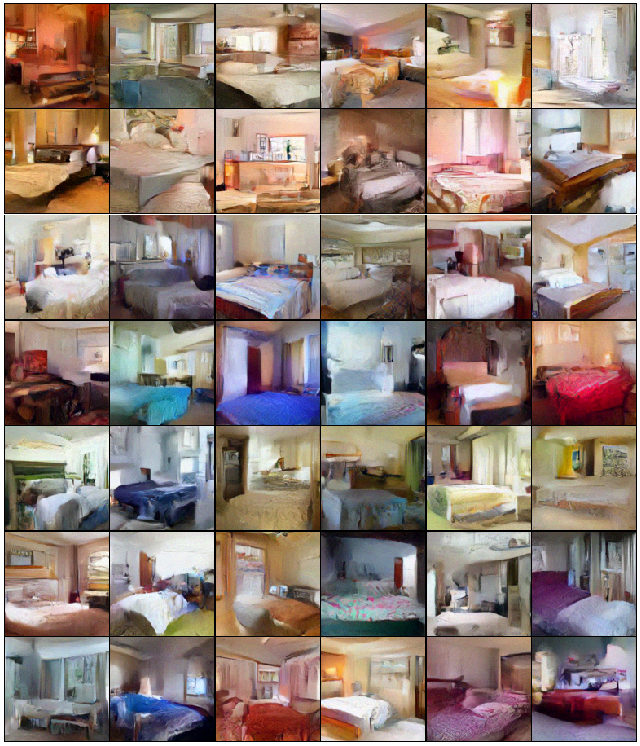}
    \caption{Samples from 5-bit, 128$\times$128 LSUN bedrooms.}
    \label{fig:bedroom}
\end{figure}

\newpage
\begin{figure}[t]
    \centering
    \includegraphics[width=\linewidth]{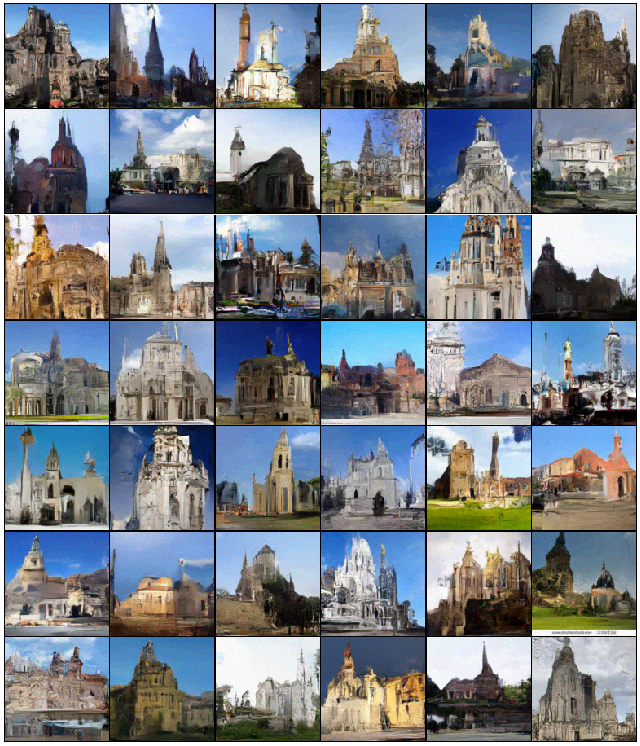}
    \caption{Samples from 5-bit, 128$\times$128 LSUN church.}
    \label{fig:church}
\end{figure}

\newpage
\begin{figure}[t]
    \centering
    \includegraphics[width=\linewidth]{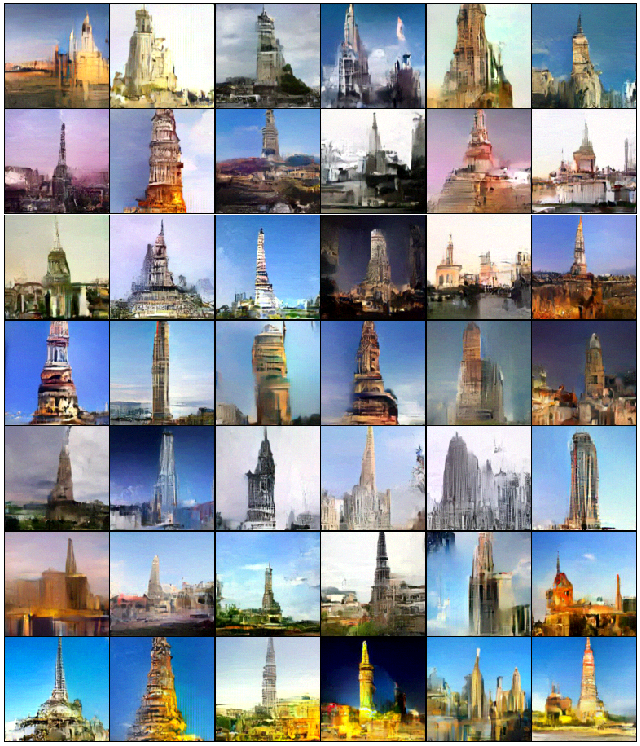}
    \caption{Samples from 5-bit, 128$\times$128 LSUN towers.}
    \label{fig:tower}
\end{figure}

\newpage
\begin{figure}[t]
    \centering
    \includegraphics[width=\linewidth]{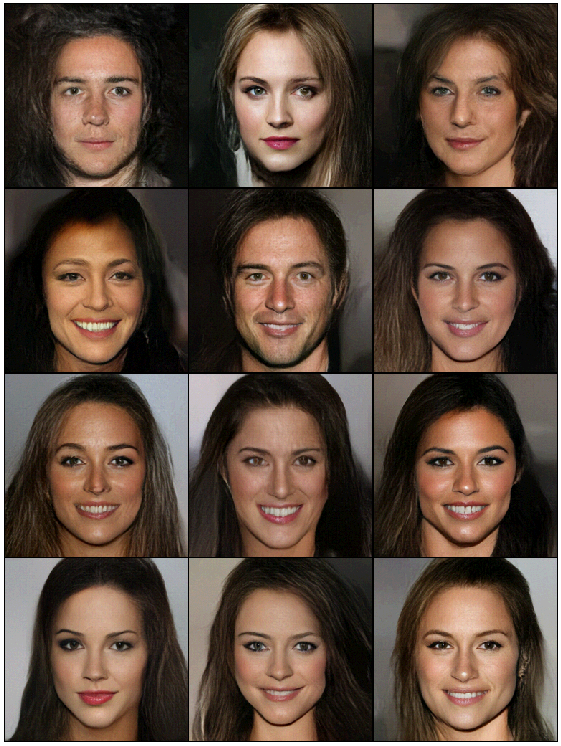}
    \caption{ Synthetic celebrities sampled from 5-bit 256$\times$256 CelebA-HQ.}
    \label{fig:celeba}
\end{figure}

\newpage
\begin{figure}[t]
    \centering
    \includegraphics[width=\textwidth]{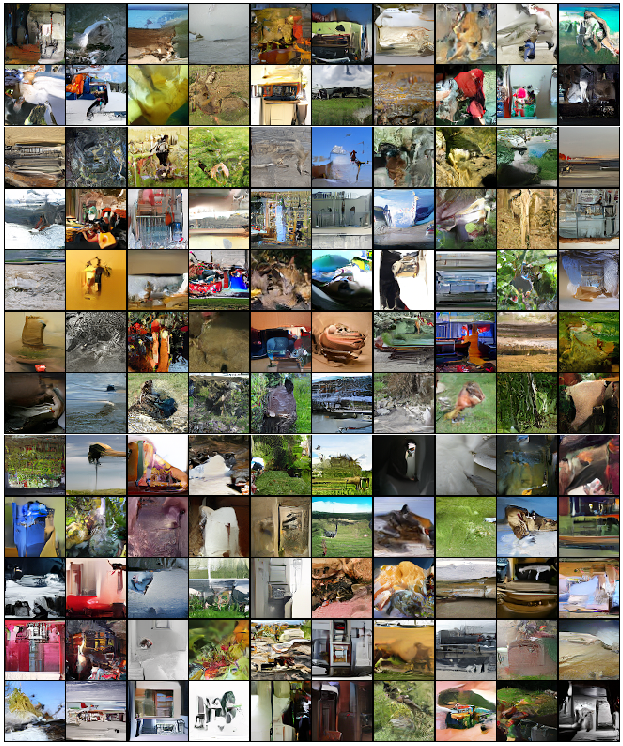}
    \caption{Samples from 8-bit imagenet 64$\times$64 with uniform  dequantization }
    \label{fig:imagenet-uni}
\end{figure}

\newpage
\begin{figure}[t]
    \centering
    \includegraphics[width=\textwidth]{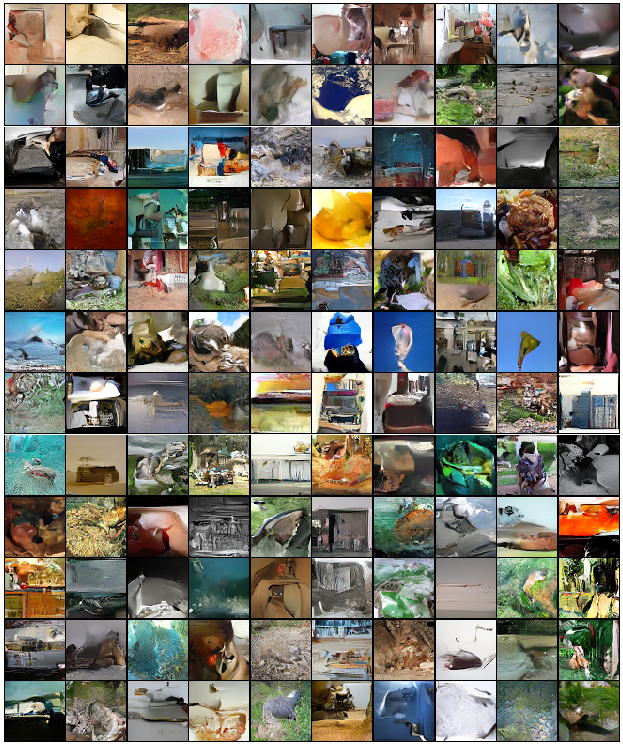}
    \caption{Samples from 8-bit imagenet 64$\times$64 with variational  dequantization}
    \label{fig:imagenet-var}
\end{figure}

\end{document}